\documentclass[11pt,a4paper]{article}
\usepackage[hyperref]{acl2020}
\usepackage{times}
\usepackage{latexsym}

\usepackage{microtype}
\usepackage{amsfonts} 
\usepackage{amsmath} 
\usepackage{multirow}
\usepackage{subfigure}
\usepackage{graphicx} 
\usepackage{booktabs} 
\usepackage{enumitem} 

\aclfinalcopy 
\def\aclpaperid{330} 

\title{DTCA: Decision Tree-based Co-Attention Networks for Explainable Claim Verification}

\author{Lianwei Wu \\
  Affiliation / Address line 1 \\
  Affiliation / Address line 2 \\
  Affiliation / Address line 3 \\
  \texttt{email@domain} \\\And
  Second Author \\
  Affiliation / Address line 1 \\
  Affiliation / Address line 2 \\
  Affiliation / Address line 3 \\
  \texttt{email@domain} \\}

\author{Lianwei Wu, Yuan Rao, Yongqiang Zhao, Hao Liang, Ambreen Nazir \\
  Lab of Social Intelligence and Complexity Data Processing, \\ School of Software Engineering, Xi'an Jiaotong University, China \\
  \textls[-30]{Shannxi Joint Key Laboratory for Artifact Intelligence\! (Sub-Lab of Xi'an Jiaotong University), China} \\
  Research Institute of Xi'an Jiaotong University, Shenzhen, China \\
  {\tt {\small \{stayhungry,yongqiang1210,favorablelearner,ambreen.nazir\}@stu.xjtu.edu.cn}} \\ {\small {\tt raoyuan@mail.xjtu.edu.cn}} \\}

\date{}

\begin{document}
\maketitle
\begin{abstract}
Recently, many methods discover effective evidence from reliable sources by appropriate neural networks for explainable claim verification, which has been widely recognized. However, in these methods, the discovery process of evidence is nontransparent and unexplained. Simultaneously, the discovered evidence only roughly aims at the interpretability of the whole sequence of claims but insufficient to focus on the false parts of claims. In this paper, we propose a Decision Tree-based Co-Attention model (DTCA) to discover evidence for explainable claim verification. Specifically, we first construct Decision Tree-based Evidence model (DTE) to select comments with high credibility as evidence in a transparent and interpretable way. Then we design Co-attention Self-attention networks (CaSa) to make the selected evidence interact with claims, which is for 1) training DTE to determine the optimal decision thresholds and obtain more powerful evidence; and 2) utilizing the evidence to find the false parts in the claim. Experiments on two public datasets, RumourEval and PHEME, demonstrate that DTCA not only provides explanations for the results of claim verification but also achieves the state-of-the-art performance, boosting the F1-score by 3.11\%, 2.41\%, respectively.
\end{abstract}

\section{Introduction}

The increasing popularity of social media has brought unprecedented challenges to the ecology of information dissemination, causing rampancy of a large volume of false or unverified claims, like extreme news, hoaxes, rumors, fake news, etc. Research indicates that during the US presidential election (2016), fake news accounts for nearly 6\% of all news consumption, where 1\% of users are exposed to 80\% of fake news, and 0.1\% of users are responsible for sharing 80\% of fake news \cite{grinberg2019fake}, and democratic elections are vulnerable to manipulation of the false or unverified claims on social media \cite{aral2019protecting}, which renders the automatic verification of claims a crucial problem.

Currently, the methods for automatic claim verification could be divided into two categories: the first is that the methods relying on deep neural networks learn credibility indicators from claim content and auxiliary relevant articles or comments (i.e., responses) \cite{volkova2017separating,rashkin2017truth,dungs2018can}. Despite their effectiveness, these methods are difficult to explain why claims are true or false in practice. To overcome the weakness, a trend in recent studies (the second category) is to endeavor to explore evidence-based verification solutions, which focuses on capturing the fragments of evidence obtained from reliable sources by appropriate neural networks \cite{popat2018declare,hanselowski2018ukp,ma2019sentence,nie2019combining}. For instance, \citet{thorne2018fever} build multi-task learning to extract evidence from Wikipedia and synthesize information from multiple documents to verify claims. \citet{popat2018declare} capture signals from external evidence articles and model joint interactions between various factors, like the context of a claim and trustworthiness of sources of related articles, for assessment of claims. \citet{ma2019sentence} propose hierarchical attention networks to learn sentence-level evidence from claims and their related articles based on coherence modeling and natural language inference for claim verification.

Although these methods provide evidence to solve the explainability of claim verification in a manner, there are still several limitations. \textbf{First}, they are generally hard to interpret the discovery process of evidence for claims, namely, lack the interpretability of methods themselves because these methods are all based on neural networks, belonging to nontransparent black box models. \textbf{Secondly}, the provided evidence only offers a coarse-grained explanation to claims. They are all aimed at the interpretability of the whole sequence of claims but insufficient to focus on the false parts of claims.

To address the above problems, we design \textbf{D}ecision \textbf{T}ree-based \textbf{C}o-\textbf{A}ttention networks (DTCA) to discover evidence for explainable claim verification, which contains two stages: 1) \textbf{D}ecision \textbf{T}ree-based \textbf{E}vidence model (DTE) for discovering evidence in a transparent and interpretable way; and 2) \textbf{C}o-\textbf{a}ttention \textbf{S}elf-\textbf{a}ttention networks (CaSa) using the evidence to explore the false parts of claims. Specifically, DTE is constructed on the basis of structured and hierarchical comments (aiming at the claim), which considers many factors as decision conditions from the perspective of content and meta data of comments and selects high credibility comments as evidence. CaSa exploits the selected evidence to interact with claims at the deep semantic level, which is for two roles: one is to train DTE to pursue the optimal decision threshold and finally obtain more powerful evidence; and another is to utilize the evidence to find the false parts in claims. Experimental results reveal that DTCA not only achieves the state-of-the-art performance but also provides the interpretability of results of claim verification and the interpretability of selection process of evidence. Our contributions are summarized as follows:

\begin{itemize}[leftmargin=*]
\item We propose a transparent and interpretable scheme that incorporates decision tree model into co-attention networks, which not only discovers evidence for explainable claim verification (Section  \ref{sec4.4.3Explainability}) but also provides interpretation for the discovery process of evidence through the decision conditions (Section \ref{sec4.4.2decision}).
\item Designed co-attention networks promote the deep semantic interaction between evidence and claims, which can train DTE to obtain more powerful evidence and effectively focus on the false parts of claims (Section \ref{sec4.4.3Explainability}).
\item Experiments on two public, widely used fake news datasets demonstrate that our DTCA achieves more excellent performance than previous state-of-the-art methods (Section \ref{sec4.3.2results}).
\end{itemize}

\section{Related Work}

\textbf{Claim Verification } Many studies on claim verification generally extract an appreciable quantity of credibility-indicative features around semantics \cite{ma2018rumor,khattar2019mvae,wu2020discovering}, emotions \cite{ajao2019sentiment}, stances \cite{ma2018detect,kochkina2018all,wu2019different}, write styles \cite{potthast2018stylometric,grondahl2019text}, and source credibility \cite{popat2018declare,baly2018predicting} from claims and relevant articles (or comments). For a concrete instance, \citet{wu2019different} devise sifted multi-task learning networks to jointly train stance detection and fake news detection tasks for effectively utilizing common features of the two tasks to improve the task performance. Despite reliable performance, these methods for claim verification are unexplainable. To address this issue, recent research concentrates on the discovery of evidence for explainable claim verification, which mainly designs different deep models to exploit semantic matching \cite{nie2019combining,zhou2019gear}, semantic conflicts \cite{baly2018integrating,dvovrak2019complexity,wu2020adaptive}, and semantic entailments \cite{hanselowski2018ukp,ma2019sentence} between claims and relevant articles. For instance, \citet{nie2019combining} develop neural semantic matching networks that encode, align, and match the semantics of two text sequences to capture evidence for verifying claims. Combined with the pros of recent studies, we exert to perceive explainable evidence through semantic interaction for claim verification.

\textbf{Explainable Machine Learning } Our work is also related to explainable machine learning, which can be generally divided into two categories: intrinsic explainability and post-hoc explainability \cite{du2018techniques}. Intrinsic explainability \cite{shu2019defend,he2015trirank,zhang2018explainable} is achieved by constructing self-explanatory models that incorporate explainability directly into their structures, which requires to build fully interpretable models for clearly expressing the explainable process. However, the current deep learning models belong to black box models, which are difficult to achieve intrinsic explainability \cite{gunning2017explainable}. Post-hoc explainability \cite{samek2017explainable,wang2018tem,chen2018visually} needs to design a second model to provide explanations for an existing model. For example, \citet{wang2018tem} combine the strengths of the embeddings-based model and the tree-based model to develop explainable recommendation, where the tree-based model obtains evidence and the embeddings-based model improves the performance of recommendation. In this paper, following the post-hoc explainability, we harness decision-tree model to explain the discovery process of evidence and design co-attention networks to boost the task performance.

\begin{figure}
	\centering
	\includegraphics[width=0.5\textwidth]{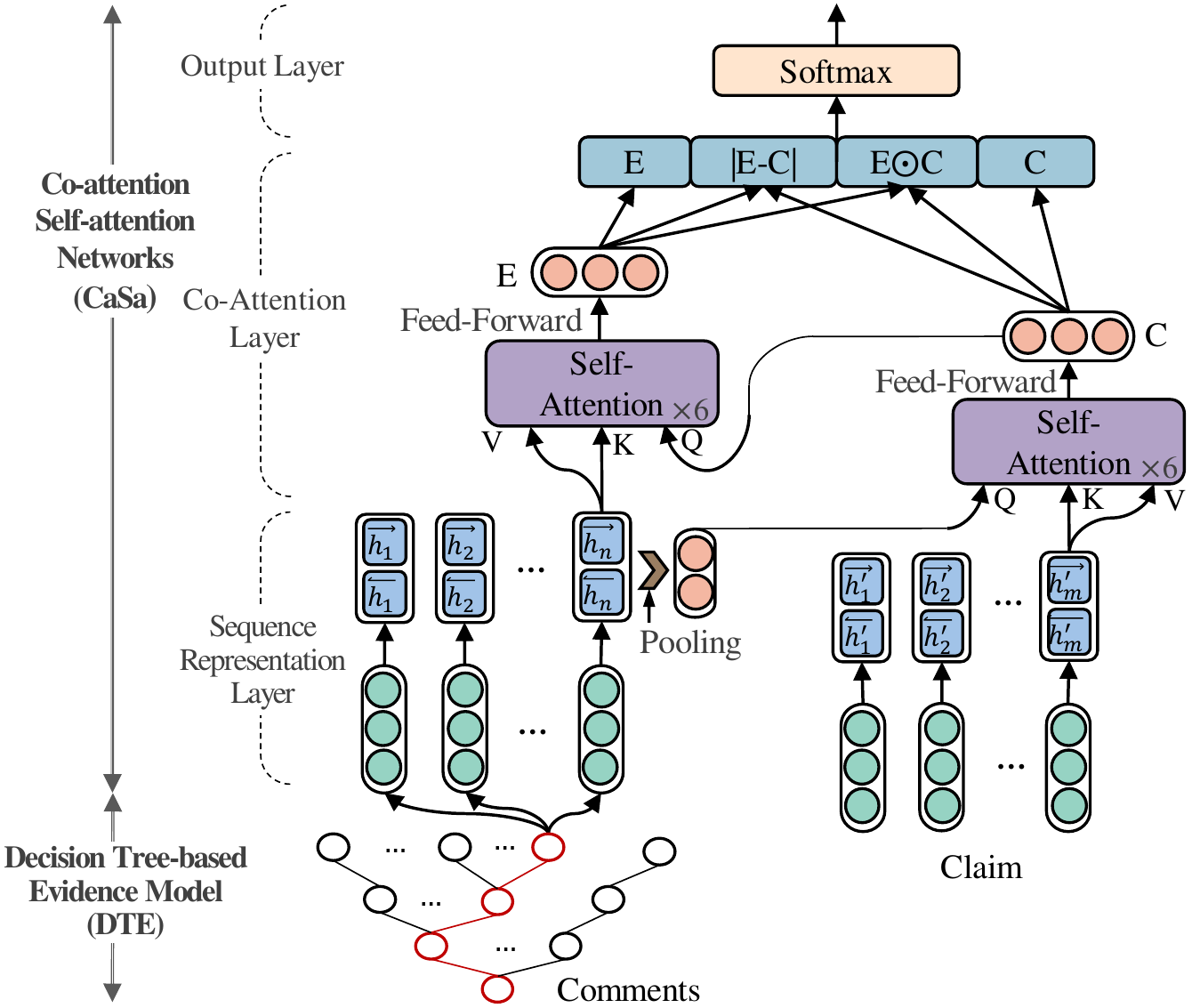}
	\caption{\textls[-10]{The architecture of DTCA. DTCA includes two stages, i.e., DTE for discovering evidence and CaSa using the evidence to explore the false parts of claims.}}
	\label{Fig1model}
\end{figure}
\section{Decision Tree-based Co-Attention Networks (DTCA)}
In this section, we introduce the decision tree-based co-attention networks (DTCA) for explainable claim verification, with architecture shown in Figure \ref{Fig1model}, which involves two stages: decision tree-based evidence model (DTE) and co-attention self-attention networks (CaSa) that consist of a 3-level hierarchical structure, i.e., sequence representation layer, co-attention layer, and output layer. Next, we describe each part of DTCA in detail.

\subsection{Decision Tree-based Evidence Model (DTE)}
DTE is based on tree comments (including replies) aiming at one claim. We first build a tree network based on hierarchical comments, as shown in the left of Figure \ref{Fig2DTE}. The root node is one claim and the second level nodes and below are users' comments on the claim ($R_{11}$, ... ,$R_{kn}$), where $k$ and $n$ denote the depth of tree comments and the width of the last level respectively. We try to select comments with high credibility as evidence of the claim, so we need to evaluate the credibility of each node (comment) in the network and decide whether to select the comment or not. Three factors from the perspective of content and meta data of comments are considered and the details are described:
\begin{figure}
	\centering
	\includegraphics[width=0.46\textwidth]{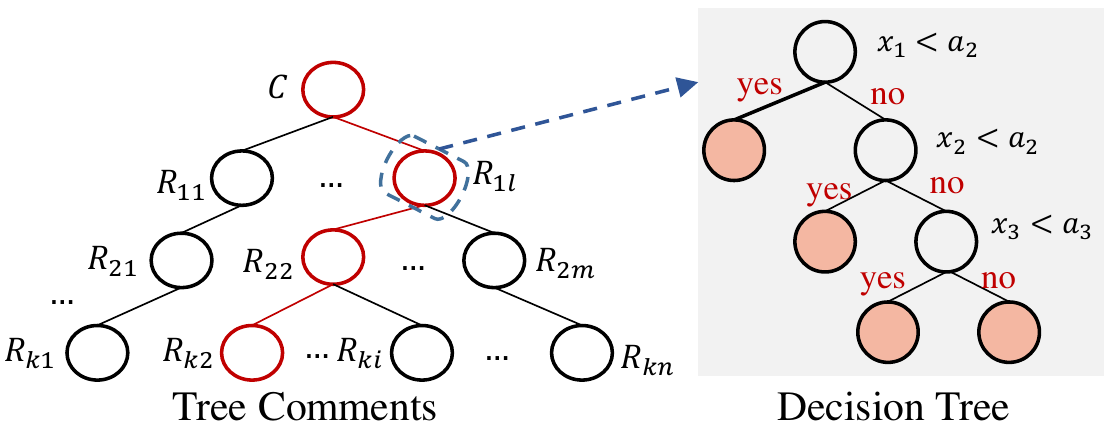}
	\caption{Overview of DTE. DTE consists of two parts: tree comment network (the left) and decision tree model (the right), which is used to evaluate the credibility of each node in the tree comment network for discovering evidence.}
	\label{Fig2DTE}
\end{figure}

\noindent \textbf{The semantic similarity between comments and claims. } It measures relevancy between comments and claims and aims to filter irrelevant and noisy comments. Specifically, we adopt soft consine measure \cite{sidorov2014soft} between average word embeddings of both claims and comments as semantic similarity.

\noindent \textbf{The credibility of reviewers\footnote{People who post comments}.} It follows that ``reviewers with high credibility also usually have high reliability in their comments" \cite{shan2016credible}. Specifically, we utilize multiple meta-data features of reviewers to evaluate reviewer credibility, i.e., whether the following elements exist or not: verified, geo, screen name, and profile image; and the number of the items: followers, friends, and favorites. The examples are shown in Appendix A.

\noindent \textbf{The credibility of comments. } It is based on meta data of comments to roughly measure the credibility of comments \cite{shu2017fake}, i.e., 1) whether the following elements exist or not: geo, source, favorite the comment; and 2) the number of favorites and content-length. The examples are shown in Appendix A.

In order to integrate these factors in a transparent and interpretable way, we build a decision tree model which takes the factors as decision conditions to measure node credibility of tree comments, as shown in the grey part in Figure \ref{Fig2DTE}.

We represent the structure of a decision tree model as $Q=\{V, E\}$, where $V$ and $E$ denote nodes and edges, respectively. Nodes in $V$ have two types: decision (a.k.a. internal) nodes and leaf nodes. Each decision node splits a decision condition $x_i$ (one of the three factors) with two decision edges (decision results) based on the specific decision threshold $a_i$. The leaf node gives the decision result (the red circle), i.e., whether the comment is selected or not. In our experiments, if any decision nodes are yes, the evaluated comment in the tree comment network will be selected as a piece of evidence. In this way, each comment is selected as evidence, which is transparent and interpretable, i.e., interpreted by decision conditions.

When comment nodes in the tree network are evaluated by the decision tree model, we leverage post-pruning algorithm to select comment subtrees as evidence set for CaSa (in section \ref{seccasa}) training.

\subsection{Co-attention Self-attention Networks (CaSa)}
\label{seccasa}
In DTE, the decision threshold $a_i$ is uncertain, to say, according to different decision thresholds, there are different numbers of comments as evidence for CaSa training. In order to train decision thresholds in DTE so as to obtain more powerful evidence, and then exploit this evidence to explore the false parts of fake news, we devise CaSa to promote the interaction between evidence and claims. The details of DTCA are as follows:

\subsubsection{Sequence Representation Layer}
The inputs of CaSa include a sequence of evidence (the evidence set obtained by DTE model is concatenated into a sequence of evidence) and a sequence of claim. Given a sequence of length $l$ tokens ${\rm \textbf{X}} = \{{\rm x}_1, {\rm x}_2,...,{\rm x}_l\}$, ${\rm \textbf{X}} \in \mathbb{R}^{l \times d}$, which could be either a claim or the evidence, each token ${\rm x}_i \in \mathbb{R}^d$ is a $d$-dimensional vector obtained by pre-trained BERT model \cite{devlin2019bert}. We encode each token into a fixed-sized hidden vector ${\rm \textbf{h}}_i$ and then obtain the sequence representation for a claim ${\rm \textbf{X}}^c$ and evidence ${\rm \textbf{X}}^e$ via two BiLSTM \cite{graves2005bidirectional} neural networks respectively.
\begin{gather}
\overrightarrow{{\rm \textbf{h}_i}} = \overrightarrow{{\rm \textbf{LSTM}}}(\overrightarrow{{\rm \textbf{h}_{i-1}}}, {\rm x}_i) \\
\overleftarrow{{\rm \textbf{h}_i}} = \overleftarrow{{\rm \textbf{LSTM}}}(\overleftarrow{{\rm \textbf{h}_{i+1}}}, {\rm x}_i)  \\
{\rm \textbf{h}_i} = [\overrightarrow{{\rm \textbf{h}_i}} ; \overleftarrow{{\rm \textbf{h}_i}}]
\end{gather}
where $\overrightarrow{{\rm \textbf{h}}_i}\!\!\in\!\! \mathbb{R}^h$ and $\overleftarrow{{\rm \textbf{h}}_i}\!\!\in\!\! \mathbb{R}^h$ are hidden states of forward LSTM $\overrightarrow{{\rm \textbf{LSTM}}}$ and backward LSTM $\overleftarrow{{\rm \textbf{LSTM}}}$. $h$ is the number of hidden units of LSTM. $;$ denotes concatenation operation. Finally, ${\rm \textbf{R}}^e\!\in\! \mathbb{R}^{l \times 2h}$ and ${\rm \textbf{R}}^c \!\in\! \mathbb{R}^{l \times 2h}$ are representations of sequences of both evidence and a claim. Additionally, experiments confirm BiLSTM in CaSa can be replaced by BiGRU \cite{cho2014properties} for comparable performance.

\subsubsection{Co-attention Layer}
Co-attention networks are composed of two hierarchical self-attention networks. In our paper, the sequence of evidence first leverages one self-attention network to conduct deep semantic interaction with the claim for capturing the false parts of the claim. Then semantics of the interacted claim focus on semantics of the sequence of evidence via another self-attention network for concentrating on the key parts of the evidence. The two self-attention networks are both based on the multi-head attention mechanism \cite{vaswani2017attention}. Given a matrix of $l$ query vectors ${\rm \textbf{Q}} \!\in\! \mathbb{R}^{l \times 2h}$, keys ${\rm \textbf{K}} \!\in\! \mathbb{R}^{l \times 2h}$, and values ${\rm \textbf{V}} \!\in\! \mathbb{R}^{l\times 2h}$, the scaled dot-product attention, the core of self-attention networks, is described as
\begin{equation}\label{eq4}
\setlength{\abovedisplayskip}{0pt}
\setlength{\belowdisplayskip}{0pt}
{\rm \textbf{Attention}}({\rm \textbf{Q}}, {\rm \textbf{K}}, {\rm \textbf{V}}) = {\rm \textbf{softmax}}(\frac{{\rm \textbf{Q}}{\rm \textbf{K}}^T}{\sqrt{d}}){\rm \textbf{V}}
\end{equation}

Particularly, to enable claim and evidence to interact more directly and effectively, in the first self-attention network, ${\rm \textbf{Q}} = {\rm \textbf{R}}_{pool}^e$ ($R_{pool}^e \in \mathbb{R}^{2h}$) is the max-pooled vector of the sequence representation of evidence, and ${\rm \textbf{K}} \!=\! {\rm \textbf{V}} \!=\! {\rm \textbf{R}}^c$, ${\rm \textbf{R}}^c$ is the sequence representation of claim. In the second self-attention network, ${\rm \textbf{Q}} \!=\! {\rm \textbf{C}}$, i.e., the output vector of self-attention network for claim (the details are in Eq. \ref{eq7}), and ${\rm \textbf{K}} \!=\! {\rm \textbf{V}} \!=\!{\rm \textbf{R}}^e$, ${\rm \textbf{R}}^e$ is the sequence representation of evidence.

To get high parallelizability of attention, multi-head attention first linearly projects queries, keys, and values $j$ times by different linear projections and then $j$ projections perform the scaled dot-product attention in parallel. Finally, these results of attention are concatenated and once again projected to get the new representation. Formally, the multi-head attention can be formulated as:
\begin{align}
\setlength{\abovedisplayskip}{0pt}
\setlength{\belowdisplayskip}{0pt}
{\rm head}_i&={\rm \textbf{Attention}}({\rm \textbf{Q}}{\rm \textbf{W}}_i^Q, {\rm \textbf{K}}{\rm \textbf{W}}_i^K, {\rm \textbf{V}}{\rm \textbf{W}}_i^V) \\
{\rm \textbf{O}}^\prime &={\rm \textbf{MultiHead}}({\rm \textbf{Q}}, {\rm \textbf{K}}, {\rm \textbf{V}})  \notag \\
&=[{\rm head}_1; {\rm head}_2; ... ; {\rm head}_j]{\rm \textbf{W}}^o
\end{align}
where ${\rm \textbf{W}}_i^Q\!\in\! \mathbb{R}^{2h\times D}$, ${\rm \textbf{W}}_i^K\!\in\! \mathbb{R}^{2h\times D}$, ${\rm \textbf{W}}_i^V \!\in\! \mathbb{R}^{2h\times D}$, and ${\rm \textbf{W}}^o\!\in\! \mathbb{R}^{2h \times 2h}$ are trainable parameters and $D$ is $2h/j$.

Subsequently, co-attention networks pass a feed forward network (FFN) for adding non-linear features while scale-invariant features, which contains a single hidden layer with an ReLU.
\begin{equation}\label{eq7}
\setlength{\abovedisplayskip}{6pt}
\setlength{\belowdisplayskip}{6pt}
{\rm \textbf{O}}\!=\!{\rm \textbf{FFN}}({\rm \textbf{O}}^\prime)\!=\!{\rm \textbf{max}}(0, {\rm \textbf{O}}^\prime{\rm \textbf{W}}_1\!+\!b_1){\rm \textbf{W}}_2\!+\!{\rm \textbf{b}}_2
\end{equation}
where ${\rm \textbf{W}}_1$, ${\rm \textbf{W}}_2$, ${\rm \textbf{b}}_1$, and ${\rm \textbf{b}}_2$ are the learned parameters. ${\rm \textbf{O}}\! =\! {\rm \textbf{C}}$ and ${\rm \textbf{O}} \!=\! {\rm \textbf{E}}$ are output vectors of two self-attention networks aiming at the claim and the evidence, respectively.

Finally, to fully integrate evidence and claim, we adopt the absolute difference and element-wise product to fuse the vectors ${\rm \textbf{E}}$ and ${\rm \textbf{C}}$ \cite{wu2019different}.
\begin{equation}\label{eq8}
\setlength{\abovedisplayskip}{1pt}
\setlength{\belowdisplayskip}{1pt}
{\rm \textbf{EC}} = [{\rm \textbf{E}};|{\rm \textbf{E}}-{\rm \textbf{C}}|;{\rm \textbf{E}} \odot {\rm \textbf{C}}; {\rm \textbf{C}}]
\end{equation}
where $\odot$ denotes element-wise multiplication operation.

\subsubsection{Output Layer}
As the last layer, softmax function emits the prediction of probability distribution by the equation:
\begin{equation}\label{eq9}
\setlength{\abovedisplayskip}{0pt}
\setlength{\belowdisplayskip}{0pt}
{\rm \textbf{p}} = {\rm \textbf{softmax}}({\rm \textbf{W}}_p {\rm \textbf{EC}} + {\rm \textbf{b}}_p)
\end{equation}

\textls[-15]{We train the model to minimize cross-entropy error for a training sample with ground-truth label ${\rm \textbf{y}}$:}
\begin{equation}\label{eq10}
\setlength{\abovedisplayskip}{0pt}
\setlength{\belowdisplayskip}{0pt}
{\rm \textbf{Loss}}=-\sum {\rm \textbf{y}}log{\rm \textbf{p}}
\end{equation}

The training process of DTCA is presented in Algorithm 1 of Appendix B.

\section{Experiments}
As the key contribution of this work is to verify claims accurately and offer evidence as explanations, we design experiments to answer the following questions:

\begin{itemize}[leftmargin=*]
\item \textbf{RQ1:} Can DTCA achieve better performance compared with the state-of-the-art models?
\item \textbf{RQ2:} How do decision conditions in the decision tree affect model performance (to say, the interpretability of evidence selection process)?
\item \textbf{RQ3:} \textls[-30]{Can DTCA make verification results easy-to-interpret by evidence and find false parts of claims?}
\end{itemize}

\subsection{Datasets}
To evaluate our proposed model, we use two widely used datasets, i.e., RumourEval \cite{derczynski2017semeval} and PHEME \cite{zubiaga2016analysing}. \textbf{Structure.} Both datasets respectively contain 325 and 6,425 Twitter conversation threads associated with different newsworthy events like Charlie Hebdo, the shooting in Ottawa, etc. A thread consists of a claim and a tree of comments (a.k.a. responses) expressing their opinion towards the claim. \textbf{Labels.} Both datasets have the same labels, i.e., true, false, and unverified. Since our goal is to verify whether a claim is true or false, we filter out unverified tweets. Table \ref{tab1dataset} gives statistics of the two datasets.

In consideration of the imbalance label distributions, besides accuracy (A), we add precision (P), recall (R) and F1-score (F1) as evaluation metrics for DTCA and baselines. We divide the two datasets into training, validation, and testing subsets with proportion of 70\%, 10\%, and 20\% respectively.

\begin{table}
\small
	\center
 \setlength{\tabcolsep}{0.6mm}{
	\begin{tabular}{|c|c|c|c|c|c|} \hline
		Subset & Veracity &\multicolumn{2}{c|}{RumourEval} &\multicolumn{2}{c|}{PHEME} \\ \cline{3-4} \cline{5-6}
		& & \#posts & \#comments & \#posts & \#comments \\ \hline  \hline
		\multirow{3}*{Training} & True & 83 & 1,949 & 861 & 24,438 \\
		& False & 70 & 1,504 & 625 & 17,676 \\
		& Total & 153 & 3,453 & 1,468 & 42,114 \\  \hline
		\multirow{3}*{Validation} & True & 10 & 101 & 95 & 1,154 \\
		& False & 12 & 141 & 115 & 1,611 \\
		& Total & 22 & 242 & 210 & 2,765 \\ \hline
		\multirow{3}*{Testing} & True & 9 & 412 & 198 & 3,077 \\
		& False & 12 & 437 & 219 & 3,265 \\
		& Total & 21 & 849 & 417 & 6,342 \\ \hline
	\end{tabular}
	}
	\caption{Statistics of the datasets.}
	\label{tab1dataset}
\end{table}

\subsection{Settings}
We turn all hyper-parameters on the validation set and achieve the best performance via a small grid search. For hyper-parameter configurations, (1) in DTE, the change range of semantic similarity, the credibility of reviewers, and the credibility of comments respectively belong to [0, 0.8], [0, 0.8], and [0, 0.7]; (2) in CaSa, word embedding size $d$ is set to 768; the size of LSTM hidden states $h$ is 120; attention heads and blocks are 6 and 4 respectively; the dropout of multi-head attention is set to 0.8; the initial learning rate is set to 0.001; the dropout rate is 0.5; and the mini-batch size is 64.

\begin{table*}
\small
	\centering
 \setlength{\tabcolsep}{1.6mm}{
	\begin{tabular}{|l|l||c|c|c|c|c|c|c|c|c|}
		\hline
		Dataset & Measure & SVM & CNN & TE & DeClarE & TRNN & MTL-LSTM & Bayesian-DL & Sifted-MTL & Ours \\ \hline \hline
		\multirow{4}*{RumourEval} & A (\%) & 71.42 & 61.90 & 66.67 & 66.67 & 76.19 & 66.67 & 80.95 & 81.48 & \textbf{82.54} \\
		& P (\%) & 66.67 & 54.54 & 60.00 & 58.33 & 70.00 & 57.14 & 77.78 & 72.24 & \textbf{78.25} \\
		& R (\%) & 66.67 & 66.67 & 66.67 & 77.78 & 77.78 & \textbf{88.89} & 77.78 & 86.31 & 85.60 \\
		& F1 (\%) & 66.67 & 59.88 & 63.15 & 66.67 & 73.68 & 69.57 & 77.78 & 78.65 & \textbf{81.76} \\\hline
		\multirow{4}*{PHEME} & A (\%) & 72.18 & 59.23 & 65.22 & 67.87 & 78.65 & 74.94 & 80.33 & 81.27 & \textbf{82.46} \\
		& P (\%) & 78.80 & 56.14 & 63.05 & 64.68 & 77.11 & 68.77 & 78.29 & 73.41 & \textbf{79.08} \\
		& R (\%) & 75.75 & 64.64 & 64.64 & 71.21 & 78.28 & 87.87 & 79.29 & \textbf{88.10} & 86.24 \\
		& F1 (\%) & 72.10 & 60.09 & 63.83 & 67.89 & 77.69 & 77.15 & 78.78 & 80.09 & \textbf{82.50} \\
		\hline
	\end{tabular}
}
	\caption{The performance comparison of DTCA against the baselines.}
	\label{Tab2performEval}
\end{table*}

\subsection{Performance Evaluation (RQ1)}
\label{sec4.3performEval}
\subsubsection{Baselines}
\textbf{SVM} \cite{derczynski2017semeval} is used to detect fake news based on manually extracted features.

\textbf{CNN} \cite{chen2017ikm} adopts different window sizes to obtain semantic features similar to n-grams for rumor classification.

\textbf{TE} \cite{guacho2018semi} creates article-by-article graphs relying on tensor decomposition with deriving article embeddings for rumor detection.

\textbf{DeClarE} \cite{popat2018declare} presents attention networks to aggregate signals from external evidence articles for claim verification.

\textbf{TRNN} \cite{ma2018rumor} proposes two tree-structured RNN models based on top-down and down-top integrating semantics of structure and content to detect rumors. In this work, we adopt the top-down model with better results as the baseline.

\textbf{MTL-LSTM} \cite{kochkina2018all} jointly trains rumor detection, claim verification, and stance detection tasks, and learns correlations among these tasks for task learning.

\textbf{Bayesian-DL} \cite{zhang2019reply} uses Bayesian to represent the uncertainty of prediction of the veracity of claims and then encodes responses to update the posterior representations.

\textbf{Sifted-MTL} \cite{wu2019different} is a sifted multi-task learning model that trains jointly fake news detection and stance detection tasks and adopts gate and attention mechanism to screen shared features.

\subsubsection{Results of Comparison}
\label{sec4.3.2results}
Table \ref{Tab2performEval} shows the experimental results of all compared models on the two datasets. We observe that:
\begin{itemize}[leftmargin=*]
\item SVM integrating semantics from claim content and comments outperforms traditional neural networks only capturing semantics from claim content, like CNN and TE, with at least 4.75\% and 6.96\% boost in accuracy on the two datasets respectively, which indicates that semantics of comments are helpful for claim verification.
\item On the whole, most neural network models with semantic interaction between comments and claims, such as TRNN and Bayesian-DL, achieve from 4.77\% to 9.53\% improvements in accuracy on the two datasets than SVM without any interaction, which reveals the effectiveness of the interaction between comments and claims.
\item TRNN, Bayesian-DL, and DTCA enable claims and comments to interact, but the first two models get the worse performance than DTCA (at least 1.06\% and 1.19\% degradation in accuracy respectively). That is because they integrate all comments indiscriminately and might introduce some noise into their models, while DTCA picks more valuable comments by DTE.
\item Multi-task learning models, e.g. MTL-LSTM and Sifted-MTL leveraging stance features show at most 3.29\% and 1.86\% boosts in recall than DTCA on the two datasets respectively, but they also bring out noise, which achieve from 1.06\% to 21.11\% reduction than DTCA in the other three metrics. Besides, DTCA achieves 3.11\% and 2.41\% boosts than the latest baseline (sifted-MTL) in F1-score on the two datasets respectively. These elaborate the effectiveness of DTCA.
\end{itemize}

\subsection{Discussions}
\subsubsection{\textls[-30]{The impact of comments on DTCA}}
In Section \ref{sec4.3performEval}, we find that the use of comments can improve the performance of models. To further investigate the quantitative impact of comments on our model, we evaluate the performance of DTCA and CaSa with 0\%, 50\%, and 100\% comments. The experimental results are shown in Table \ref{Tab3diffComm}. We gain the following observations:
\begin{itemize}[leftmargin=*]
\item Models without comment features present the lowest performance, decreasing from 5.08\% to 9.76\% in accuracy on the two datasets, which implies that there are a large number of veracity-indicative features in comments.
\item As the proportion of comments expands, the performance of models is improved continuously. However, the rate of comments for CaSa raises from 50\% to 100\%, the boost is not significant, only achieving 1.44\% boosts in accuracy on RumourEval, while DTCA obtains better performance, reflecting 3.90\% and 3.28\% boosts in accuracy on the two datasets, which fully proves that DTCA can choose valuable comments and ignore unimportant comments with the help of DTE.
\end{itemize}

\begin{table}
\small
	\centering
	\begin{tabular}{|l|l||c|c|c|c|}
		\hline
        \multicolumn{6}{|c|}{No (0\%) Comments} \\ \hline
		 & & A & P & R & F1 \\ \hline \hline
		\multirow{2}*{RumourEval} & CaSa & 72.78 & 67.03 & 72.87 & 69.83 \\
         & DTCA & 72.78 & 67.03 & 72.87 & 69.83 \\ \hline
        \multirow{2}*{PHEME} & CaSa & 73.21 & 71.26 & 74.74 & 72.96 \\
         & DTCA & 73.21 & 71.26 & 74.74 & 72.96 \\ \hline
         \multicolumn{6}{|c|}{50\% Comments} \\ \hline
         \multirow{2}*{RumourEval} & CaSa & 76.42 & 70.21 & 76.78 & 73.35 \\
         & DTCA & 78.64 & 73.43 & 80.06 & 76.60 \\ \hline
         \multirow{2}*{PHEME} & CaSa & 77.65 & 74.18 & 78.11 & 76.09 \\
         & DTCA & 79.18 & 75.24 & 80.66 & 77.86 \\ \hline
         \multicolumn{6}{|c|}{All (100\%) Comments} \\ \hline
         \multirow{2}*{RumourEval} & CaSa & 77.86 & 71.92 & 79.24 & 75.40 \\
         & DTCA & 82.54 & 78.25 & 85.60 & 81.76 \\ \hline
         \multirow{2}*{PHEME} & CaSa & 79.85 & 75.06 & 80.35 & 77.61 \\
         & DTCA & 82.46 & 79.08 & 86.24 & 82.50 \\
		\hline
	\end{tabular}
	\caption{The performance comparison of models on different number of comments.}
	\label{Tab3diffComm}
\end{table}

\begin{figure*}
  \centering
  \subfigure[Different semantic similarity]{
    \begin{minipage}[b]{0.32\textwidth}
      \includegraphics[width=1\textwidth]{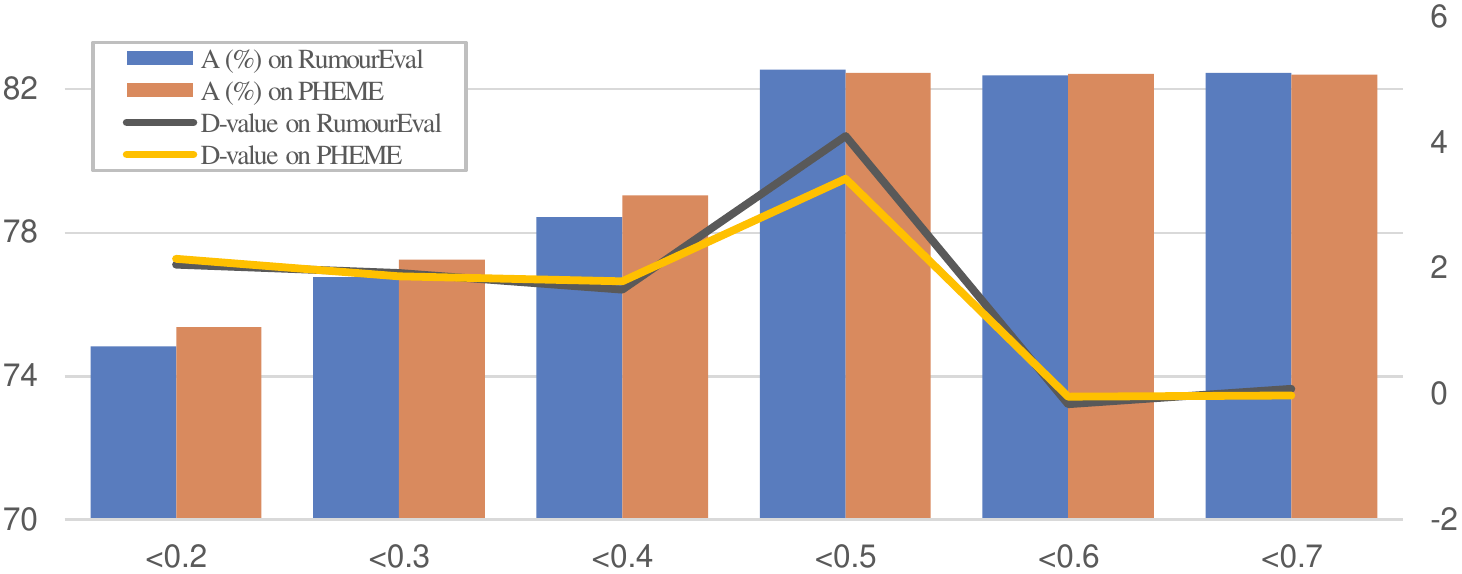}
    \end{minipage}
  }
  \subfigure[Different credibility of reviewers]{
    \begin{minipage}[b]{0.31\textwidth}
      \includegraphics[width=1\textwidth]{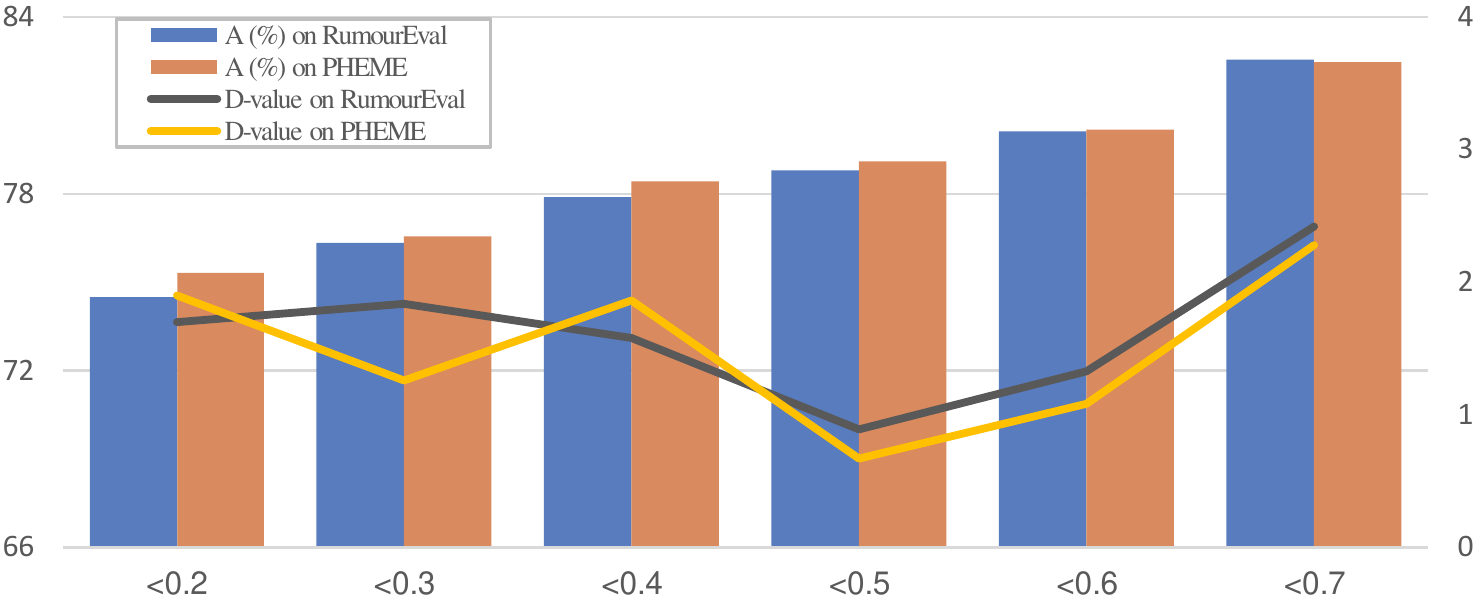}
    \end{minipage}
  }
  \subfigure[Different credibility of comments]{
    \begin{minipage}[b]{0.32\textwidth}
      \includegraphics[width=1\textwidth]{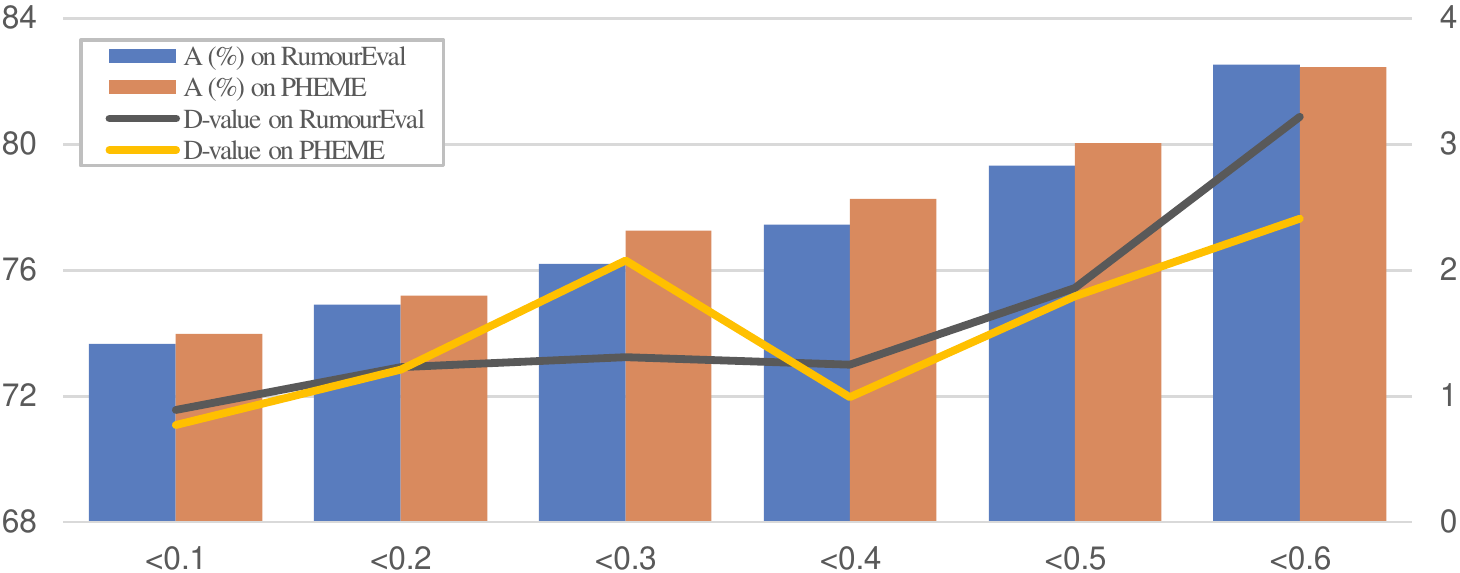}
    \end{minipage}
  }
  \caption{The accuracy comparison of DTCA in different decision conditions. Broken lines represent the performance difference (D-value) between the current decision condition and the previous decision condition.}
  \label{Fig3diffcond}
\end{figure*}
\subsubsection{The impact of decision conditions of DTE on DTCA (RQ2)}
\label{sec4.4.2decision}
To answer RQ2, we analyze the changes of model performance under different decision conditions. Different decision conditions can choose different comments as evidence to participate in the model learning. According to the performance change of the model verification, we are capable of well explaining the process of evidence selection through decision conditions. Specifically, we measure different values (interval [0, 1]) as thresholds of decision conditions so that DTE could screen different comments. Figure \ref{Fig3diffcond}(a), (b), and (c) respectively present the influence of semantic similarity ($simi$), the credibility of reviewers ($r\_cred$), and the credibility of comments ($c\_cred$) on the performance of DTCA, where the maximum thresholds are set to 0.7, 0.7, and 0.6 respectively because there are few comments when the decision threshold is greater than these values. We observe that:
\begin{itemize}[leftmargin=*]
\item When $simi$ is less than 0.4, the model is continually improved, where the average performance improvement is about 2 \% (broken lines) on the two datasets when $simi$ increases by 0.1. Especially, DTCA earns the best performance when $simi$ is set to 0.5 ($<$0.5), while it is difficult to improve performance after that. These exemplify that DTCA can provide more credibility features under appropriate semantic similarity for verification.
\item DTCA continues to improve with the increase of $r\_cred$, which is in our commonsense, i.e., the more authoritative people are, the more credible their speech is. Analogously, DTCA boosts with the increase of $c\_cred$. These show the reasonability of the terms of both the credibility of reviewers and comments built by meta data.
\item When $simi$ is set to 0.5 ($<$0.5), $r\_cred$ is 0.7 ($<$0.7), $c\_cred$ is 0.6 ($<$0.6), DTCA wins the biggest improvements, i.e., at least 3.43\%, 2.28\%, and 2.41\% on the two datasets respectively. At this moment, we infer that comments captured by the model contain the most powerful evidence for claim verification. This is, the optimal evidence is formed under the conditions of moderate semantic similarity, high reviewer credibility, and higher comment credibility, which explains the selection process of evidence.
\end{itemize}

\begin{figure*}
\centering
\includegraphics[width=1\textwidth]{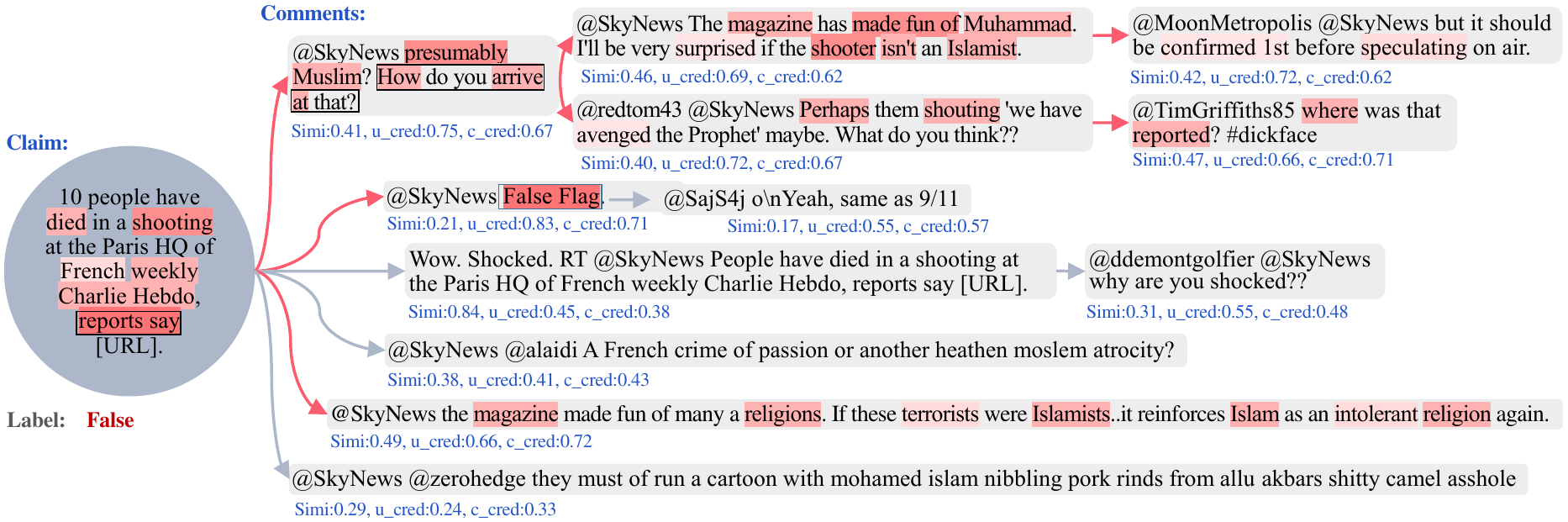}
\caption{The visualization of a sample (labeled false) in PHEME by DTCA, where the captured evidence (red arrows) and the specific values of decision conditions (blue) are presented by DTE, and the attention of different words (red shades) is obtained by CaSa.}
\label{Fig4example}
\end{figure*}

\subsubsection{Explainability Analysis (RQ3)}
\label{sec4.4.3Explainability}
To answer RQ3, we visualize comments (evidence) captured by DTE and the key semantics learned by CaSa when the training of DTCA is optimized. Figure \ref{Fig4example} depicts the results based on a specific sample in PHEME, where at the comment level, red arrows represent the captured evidence and grey arrows denote the unused comments; at the word level, darker shades indicate higher weights given to the corresponding words, representing higher attention. We observe that:
\begin{itemize}[leftmargin=*]
\item In line with the optimization of DTCA, the comments finally captured by DTE contain abundant evidence that questions the claim, like `presumably Muslim? How do you arrive at that?', `but it should be confirmed 1st before speculating on air.', and `false flag', to prove the falsity of the claim (the label of the claim is false), which indicates that DTCA can effectively discover evidence to explain the results of claim verification. Additionally, there are common characteristics in captured comments, i.e., moderate semantic similarity (interval [0.40, 0.49]), high reviewer credibility (over 0.66), and high comment credibility (over 0.62). For instance, the values of the three characteristics of evidence `@TimGriffiths85 where was that reported? \#dickface' are 0.47, 0.66, and 0.71 respectively. These phenomena explain that DTCA can give reasonable explanations to the captured evidence by decision conditions of DTE, which visually reflects the interpretability of DTCA method itself.
\item At the word level, the evidence-related words `presumably Muslim', `made fun of', `shooter', and `isn't Islamist' in comments receive higher weights than the evidence-independent words `surprised', `confirmed 1st' and `speculating', which illustrates that DTCA can earn the key semantics of evidence. Moreover, `weekly Charlie Hebdo' in the claim and `Islamist' and `Muhammad' in comments are closely focused, which is related to the background knowledge, i.e., weekly Charlie Hebdo is a French satirical comic magazine which often publishes bold satire on religion and politics. `report say' in claim is queried in the comments, like `How do you arrive at that?' and `false flag'. These visually demonstrate that DTCA can uncover the questionable and even false parts in claims.
\end{itemize}

\subsubsection{Error Analysis}
Table \ref{diffClaimComms} provides the performance of DTCA under different claims with different number of comments. We observe that DTCA achieves the satisfactory performance in claims with more than 8 comments, while in claims with less than 8 comments, DTCA does not perform well, underperforming its best performance by at least 4.92\% and 3.10\% in accuracy on the two datasets respectively. Two reasons might explain the issue: 1) The claim with few comments has limited attention, and its false parts are hard to be found by the public; 2) DTCA is capable of capturing worthwhile semantics from multiple comments, but it is not suitable for verifying claims with fewer comments.
\begin{table}
\small
	\centering
\setlength{\tabcolsep}{0.3mm}{
	\begin{tabular}{|l|l|c|c|c|c|}
		\hline
         Claims & Datasets & A (\%) & P (\%) &R (\%) & F1 (\%)  \\ \hline \hline
          \textls[-30]{Claims with less} & \textls[-30]{RumourEval} & 73.42 & 68.21 & 73.50 & 70.76 \\
          than 3 comments & PHEME & 74.10 & 72.45 & 75.32 & 73.86 \\  \hline
          \textls[-30]{Claims with com-} & \textls[-30]{RumourEval} & 75.33 & 69.16 & 75.07 & 71.99 \\
           ments $\in$ [3, 8] & PHEME & 77.26 & 74.67 & 79.03 & 76.79 \\ \hline
          \textls[-30]{Claims with more } & \textls[-30]{RumourEval} & 80.25 & 75.61 & 83.45 & 79.34\\
          \textls[-30]{than 8 comments} & PHEME & 80.36 & 75.52 & 84.33 & 79.68 \\ \hline
	\end{tabular}
}
	\caption{The performance comparison of DTCA under different claims with different number of comments.}
	\label{diffClaimComms}
\end{table}

\section{Conclusion}
We proposed a novel framework combining decision tree and neural attention networks to explore a transparent and interpretable way to discover evidence for explainable claim verification, which constructed decision tree model to select comments with high credibility as evidence, and then designed co-attention networks to make the evidence and claims interact with each other for unearthing the false parts of claims. Results on two public datasets demonstrated the effectiveness and explainability of this framework. In the future, we will extend the proposed framework by considering more context (meta data) information, such as time, storylines, and comment sentiment, to further enrich our explainability.

\section*{Acknowledgments}
The research work is supported by National Key Research and Development Program in China (2019YFB2102300); The World-Class Universities (Disciplines) and the Characteristic Development Guidance Funds for the Central Universities of China (PY3A022); Ministry of Education Fund Projects\! (18JZD022 and 2017B00030); Shenzhen Science and Technology Project (JCYJ20180306170836595); Basic Scientific Research Operating Expenses of Central Universities (ZDYF2017006); Xi'an Navinfo Corp.\& Engineering Center of Xi'an Intelligence Spatial-temporal Data Analysis Project (C2020103); Beilin District of Xi'an Science \& Technology Project (GX1803).
We would like to thank the anonymous reviewers for their insightful comments.

\bibliography{references-acl2020}
\bibliographystyle{acl_natbib}

\clearpage
\appendix

\section{The Details of Decision Conditions}
\label{sec:appendix}
The paper has introduced the details of semantic similarity between claims and comments, here we introduce the details of the other two decision conditions.

Table \ref{tab4rcred} shows some scores of meta data related to reviewer credibility. In elements `whether the elements exist or not', if the element is false, the score will be zero. The credibility score of reviewer ($r\_cred$) is formulated as follows:
\begin{equation}\label{eqrcred}
r\_cred = \frac{A1}{B1}
\end{equation}
where A1 denotes the specific score accumulation of all metadata related to reviewer credibility and B1 means the total credibility score of reviewers.

Table \ref{tab5ccred} describes some credibility score of comments. Like the credibility score of reviewer, in elements `whether the elements exist or not', if the element is false, the score will be zero. The credibility score of comments ($c\_cred$) is formulated as follows:
\begin{equation}\label{eqccred}
c\_cred = \frac{A2}{B2}
\end{equation}
where A2 denotes the specific score accumulation of all metadata related to comment credibility and B2 means the total credibility score of comments.

\begin{table}
\small
	\centering
\setlength{\tabcolsep}{0.3mm}{
	\begin{tabular}{|l||l|c|c|}
		\hline
         & \multirow{2}*{Element} & Standard of & \multirow{2}*{Credibility score} \\
         & & criterion &  \\ \hline
         Whether the & verified & True & 3 \\
         elements exist & geo & True & 3 \\
         or not & screen name & True & 1 \\
          & profile image & True & 2 \\ \hline
         The value of & \multirow{3}*{followers} & [0, 100) & 2 \\
         elements &  & [100, 500) & 5 \\
          &  & [500, $\infty$) & 10 \\ \cline{2-4}
          &friends& [0, 100) & 1 \\
          & & [100, 200) & 2 \\
          & & [200, $\infty$) & 5 \\ \cline{2-4}
          &favourites& [0, 100) & 2 \\
          & & [100, 200) & 3 \\
          & & [200, $\infty$) & 5 \\ \hline
	\end{tabular}
}
	\caption{The credibility score of reviewers}
	\label{tab4rcred}
\end{table}

\begin{table}
\small
	\centering
\setlength{\tabcolsep}{0.3mm}{
	\begin{tabular}{|l||l|c|c|}
		\hline
         & \multirow{2}*{Element} & Standard of & Credibility \\
         & & criterion & score \\ \hline
         \multirow{2}*{Whether the elements} & geo & True & 3 \\
         & source & True & 3 \\  \cline{2-4}
         \multirow{2}*{exist or not} & favorite & \multirow{2}*{True} & \multirow{2}*{1} \\
         & the comment & & \\ \hline
         \multirow{4}*{The value of elements} & \multirow{2}*{the number of favorites} & [0, 100) & 5 \\
         &  & [100, $\infty$) & 7 \\ \cline{2-4}
          & \multirow{2}*{the length of content} & [0, 10) & 3 \\
          &  & [10, $\infty$) & 7 \\ \hline
	\end{tabular}
}
	\caption{The credibility score of comments}
	\label{tab5ccred}
\end{table}

\section{Algorithm of DTCA}
\label{algorithm}
The training procedure of DTCA is shown in Algorithm 1.
\begin{table}
\small
	\center
	\begin{tabular}{rl} \toprule
    \multicolumn{2}{l}{\textbf{Algorithm 1:} Training Procedure of DTCA.} \\ \hline
    \multicolumn{2}{l}{\textbf{Require: } Dataset $S=\{C_i,R_i,y\}_1^T$ with $T$ training samples,} \\
    \multicolumn{2}{l}{where $C_i$ denotes one claim and $R_i$ means tree comments,} \\
    \multicolumn{2}{l}{ i.e., $R_i={r_1, r_2, ..., r_k}$. Particularly, $k$ is different for }\\
    \multicolumn{2}{l}{ different claims; the thresholds of the three decision conditions }\\
    \multicolumn{2}{l}{ are $a_1$, $a_2$, $a_3$, respectively; } \\
    \multicolumn{2}{l}{the evidence set $E$; model parameters $\Theta$; learning rate $\epsilon$. } \\
    \scriptsize\textbf{1}& \textbf{Initial parameters;} \\
    \scriptsize\textbf{2}& \textbf{Repeat} \\
    \scriptsize\textbf{3}& $\big|$ \quad \textbf{For} $i=1$ to $T$ \textbf{do} \\
    & $\big|$ \quad $\big|$\quad // Part 1: DTE \\
    \scriptsize\textbf{4}& $\big|$ \quad $\big|$\quad A series of subtree sets S obtained by depth-first \\
    &$\big|$ \quad $\big|$\quad search of tree comments $R_i$, i.e., $S\!=\![S_1,\!S_2, ...,\!S_n]$;\\
    \scriptsize\textbf{5}& $\big|$ \quad $\big|$\quad On each subtree $S_i$: \\
    \scriptsize\textbf{6}& $\big|$ \quad $\big|$\quad \textbf{For} $r_i$ in $S_i$ \textbf{do} \\
    \scriptsize\textbf{7}& $\big|$ \quad $\big|$\quad $\big|$\quad // The semantic similarity between comments \\
    & $\big|$ \quad $\big|$\quad $\big|$\quad // and claims \\
    \scriptsize\textbf{8}& $\big|$ \quad $\big|$\quad $\big|$\quad \textbf{If} $simi(r_i, C_i)\in interval[a, b]$: \\
    \scriptsize\textbf{9}& $\big|$ \quad $\big|$\quad $\big|$\quad\quad $E = E + r_i$; \\
    \scriptsize\textbf{10}& $\big|$ \quad $\big|$\quad $\big|$\quad // The credibility of reviewers \\
    \scriptsize\textbf{11}& $\big|$ \quad $\big|$\quad $\big|$\quad \textbf{If} $r\_cred(r_i) > a_2$: \\
    \scriptsize\textbf{12}& $\big|$ \quad $\big|$\quad $\big|$\quad\quad $E = E + r_i$; \\
    \scriptsize\textbf{13}& $\big|$ \quad $\big|$\quad $\big|$\quad // The credibility of comments \\
    \scriptsize\textbf{14}& $\big|$ \quad $\big|$\quad $\big|$\quad \textbf{If} $c\_cred(r_i)>a_3$: \\
    \scriptsize\textbf{15}& $\big|$ \quad $\big|$\quad $\big|$\quad\quad $E = E + r_i$; \\
    \scriptsize\textbf{16}& $\big|$ \quad $\big|$\quad \textbf{End For} \\
    \scriptsize\textbf{17}& $\big|$ \quad $\big|$\quad By traversing all subtrees, the final evidence set $E$ \\
    & $\big|$ \quad $\big|$\quad that meets the conditions is captured. \\

    \scriptsize\textbf{18}& $\big|$ \quad $\big|$\quad //Part 2: CaSa \\
    \scriptsize\textbf{19}& $\big|$ \quad $\big|$\quad Word embeddings of evidence $E$: \\
    & $\big|$ \quad $\big|$\quad $X^e = BERT\_embed(E)$; \\
    \scriptsize\textbf{20}& $\big|$ \quad $\big|$\quad Word embeddings of claim $C_i$: $BERT\_embed(C_i)$; \\
    \scriptsize\textbf{21}& $\big|$ \quad $\big|$\quad Get representations of evidence and claim by \\
    & $\big|$ \quad $\big|$\quad Eq. (1-3), i.e., $R^e$ and $R^c$; \\
    \scriptsize\textbf{22}& $\big|$ \quad $\big|$\quad Get $R_{pool}^e$ by maximum pooling operation of $R^e$; \\
    \scriptsize\textbf{23}& $\big|$ \quad $\big|$\quad Get deep interaction semantics $C$ of claim concerned \\
    & $\big|$ \quad $\big|$\quad  by evidence through integrating $R_{pool}^e$ into \\
    & $\big|$ \quad $\big|$\quad self-attention networks by Eq. (4-7); \\
    \scriptsize\textbf{24}& $\big|$ \quad $\big|$\quad Get deep interaction semantics $E$ of evidence \\
    & $\big|$ \quad $\big|$\quad concerned by the claim through integrating $C$ into \\
    \scriptsize\textbf{}& $\big|$ \quad $\big|$\quad  self-attention networks by Eq. (4-7); \\
    \scriptsize\textbf{25}& $\big|$ \quad $\big|$\quad Get fused vectors between evidence and claim \\
    & $\big|$ \quad $\big|$\quad by Eq. (8); \\
    \scriptsize\textbf{26}& $\big|$ \quad $\big|$\quad Compute loss $L(\Theta)$ using Eq. (9,10); \\
    \scriptsize\textbf{27}& $\big|$ \quad $\big|$\quad Compute gradient $\nabla(\Theta)$; \\
    \scriptsize\textbf{28}& $\big|$ \quad $\big|$\quad Update model: $\Theta \leftarrow \Theta - \epsilon \nabla(\Theta)$; \\
    \scriptsize\textbf{29}& $\big|$ \quad \textbf{End For} \\
    \scriptsize\textbf{30}& Update parameters $a_1$, $a_2$, $a_3$; \\
    \scriptsize\textbf{31}& Until $a_1=0.8$, $a_2=0.8$, and $a_3=0.7$. \\
    \bottomrule
	\end{tabular}
\end{table}

\end{document}